\documentclass[article]{elsarticle}
\usepackage{hyperref}
\usepackage{mathtools}
\usepackage{color}
\biboptions{numbers,sort&compress}
\usepackage[none]{hyphenat}
\usepackage[section]{placeins}

\begin{document}
\begin{frontmatter}

\title{Broad learning system with Takagi-Sugeno fuzzy subsystem for tobacco origin identification based on near infrared spectroscopy}
\author[address1]{Di Wang}
\author[address2]{Simon X. Yang}

\address[address1]{School of Information Science and Engineering, Chongqing Jiaotong University, Chongqing 400074, China}
\address[address2]{School of Engineering, University of Guelph, Guelph, Ontario, N1G 2W1, Canada\\ e-mail of corresponding author: syang@uoguelph.ca}

\begin{abstract}
\sloppy{}
Tobacco origin identification is significantly important in tobacco industry. Modeling analysis for sensor data with near infrared spectroscopy has become a popular method for rapid detection of internal features. However, for sensor data analysis using traditional artificial neural network or deep network models, the training process is extremely time-consuming. In this paper, a novel broad learning system with Takagi-Sugeno (TS) fuzzy subsystem is proposed for rapid identification of tobacco origin. Incremental learning is employed in the proposed method, which obtains the weight matrix of the network after a very small amount of computation, resulting in much shorter training time for the model, with only about 3 seconds for the extra step training. The experimental results show that the TS fuzzy subsystem can extract features from the near infrared data and effectively improve the recognition performance. The proposed method can achieve the highest prediction accuracy (95.59$\%$) in comparison to the traditional classification algorithms, artificial neural network, and deep convolutional neural network, and has a great advantage in the training time with only about 128 seconds.

\end{abstract}


\end{frontmatter}

\section{Introduction}\label{sec:introduction}
\sloppy{}
Although tobacco is harmful to human health, research on tobacco still has certain value and significance. Tobacco has been widely known as a commercial crop, and the quality and flavor of tobacco leaf are affected heavily by the cultivation geographical region \cite{wd_2019_lar,wd_2017}. Therefore, distinguishing the growing regions of tobacco leaves plays a significant role before putting them into products \cite{tobacco_1,tobacco_2,tobacco_3}. In practical application in tobacco industry, growing region evaluation is usually conducted by some experts according to the sensory inspection like sense of smell, taste, and smoke quality, which is very time-consuming, low efficiency and laborious. The manual evaluation of growing region depends on the experiences of experts to a great extent, and the results could be affected by the experts’ emotional state as well as the environment outside such as the light and temperature of the assessment place. Such kind of manual evaluation of growing region is not reliable and cannot meet the requirement of reproducible assessment process for the tobacco quality control and supervision. 

With the development of soft computing technology, many identification techniques for tobacco origin have been applied. As the distribution of metal elements in tobacco leaves from different producing areas is different, X-ray fluorescence spectroscopy was used to rapidly detect the distribution of various metal elements in tobacco leaves, and discriminant analysis was combined to identify the tobacco origin \cite{c1}. Electronic nose was adopted to collect data from different sensors and combined principal component analysis, clustering analysis, and other algorithms to identify tobacco origin \cite{c2}. It has proven that pollen analysis is able to assist with identifying geographical origin of tobacco \cite{c3}. As chemical composition content being treated as independent variable affecting the producing area of flue-cured tobacco, Naive Bayes classification algorithm was established for pattern recognition of tobacco origin \cite{c4}.

Near infrared (NIR) spectroscopy has been known as a kind of fast, nondestructive and accurate data analysis technology for origin identification in recent years \cite{c5,c6,c7,c8li2020near,c9,c10liu2019rapid,wd_svm}. NIR spectroscopy combined with other soft computing methods has been widely applied for tobacco origin identification. Zhu et al. used high dimensional feature grouping method in NIR spectra to identify tobacco growing area \cite{c11}. Wang et al. collected twelve hundred seventy six superior tobacco leaf from four producing areas, which are Yuxi, Chuxiong, Zhaotong and Dali in Yunnan province, and applied NIR spectrum projection and correlation methods for tobacco quality analysis of different producing areas \cite{c12}.

Although the above soft computing methods obtain better effect than manual sensory inspection for identifying tobacco growing regions, there is room for improvement in recognition performance.

Deep learning for deep structural neural network is a popular machine learning algorithm in recent years \cite{c13}. It has achieved breakthrough success in many fields \cite{c13,c14,c15,c16,c16_1}, especially in big data mining. Lu et al. conducted classification modeling method for NIR spectroscopy of tobacco leaves based on Convolution Neural Network (CNN) \cite{c16meng2018study}. Although the deep network shows amazing effects and strong prediction performance, due to the coupling between layers requires a large number of hyper-parameters, the complexity not only increases the difficulty of analyzing the deep network structure, but also consumes a lot of time and resources in the training process \cite{c17lee2018deep}. Wang et al. \cite{c18wang2020improved} proposed CNN to discriminate the tobacco cultivation regions and achieves the prediction precision with 93.16$\%$, but takes about 6833 seconds for the training process. 

In addition, with the arrival of the era of big data, the sensor data acquisition quantity increases quickly. When the components or property of the new data are changed, the analysis of new data  with the trained model before will be affected. New data and the previous data need to be retrained for establishing mathematical model, which is extremely time-consuming and inconvenient.

Therefore, to address the time-consuming problems happened in deep network during training for big data and retraining the network due to new sensor data, the tobacco origin classification model of fuzzy broad learning system based on Takagi-Sugeno (FBLS-TS) is proposed in this paper. In this soft computing method, the effective features are extracted adaptively through TS fuzzy subsystem, and the pseudo-inverse calculation and incremental learning method are applied for data processing to improve the speed of model training.

\section{Materials and Methods}\label{sec:relatedword}
\sloppy{}
In this section, the FBLS-TS model and data source used in this study are first introduced. After that, the model evaluation criteria are presented in.

\subsection{Broad learning system}
\begin{figure}[ht]
    \centering
    \includegraphics[width=12cm]{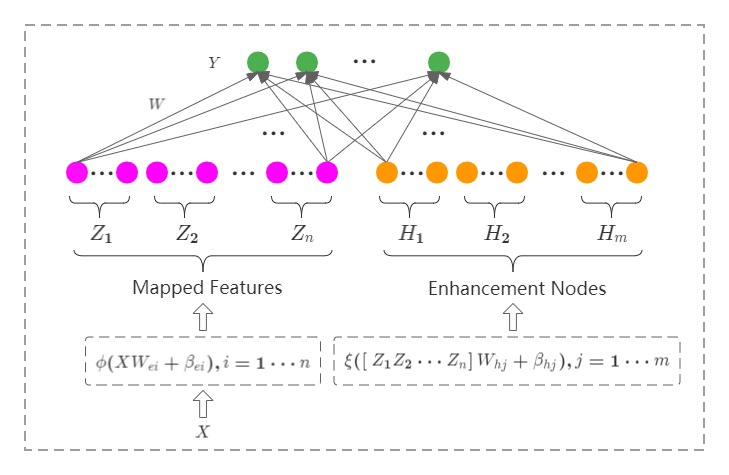}
    \caption{Network structure of BLS.}   
    \label{BLS}
\end{figure}

Broad Learning System (BLS) is an efficient incremental learning system without deep structures, which is proposed by Professor Chen of the University from Macau university in 2018 \cite{c19chen2017broad}. Since there are no multi-layer connections, BLS does not need to use gradient descent to update the weights, so the calculation speed is much faster than that of deep learning algorithms \cite{c20chen2018universal,c21luo2018intelligent}. The principle of the BLS is as follows, it firstly extracts feature from raw sensor data through linear transformation, and makes the extracted features to be feature nodes; then the feature nodes are enhanced to be enhanced nodes; lastly, both of the feature nodes and the enhanced nodes are taken as the input layer and directly connected to the output layer, and the connection weight from the input to the output is obtained through the pseudo-inverse of the network input. Supposing that $A$ is the input matrix which consists of feature nodes and enhanced nodes, $Y$ is the output, and $W$ is the weight matrix of the network, the objective of the training is to find the connection weight, which is given as 

\begin{equation}
\begin{aligned}
W\text{=}{{A}^{\text{-1}}}Y,
\end{aligned}
\end{equation}
where ${{A}^{\text{-1}}}$ is the inverse matrix of the input for the entire network, and $Y$ is the label of data. The network structure of the BLS is shown in Figure \ref{BLS}. In the figure, $X$ is the input of the network, $Z_1$, $Z_2$, …, $Z_n$ are the feature nodes, $H_1$, $H_2$, …, $H_m$ are the enhanced nodes, $W$ is the weight of network and $Y$ is the output.
\subsubsection{Feature nodes}
Assuming the NIR data of tobacco is $X_{s\times f}^{'}$, where $s$ is the number of training sample, and $f$ is the number of features. Normalization is conducted on $X_{s\times f}^{'}$ to ensure that the data was normalized within the same range firstly, and then a column of 1 is added to the last column of $X_{s\times f}^{'}$ to make it to be ${{X}_{s\times (f+1)}}$, the purpose of which is to make it convenient to add the bias in the matrix calculation when generating feature nodes. Supposing that ${{N}_{1}}$ is the number of feature nodes in per window of the network structure. Assuming ${{N}_{2}}$ is the number of windows of the network structure and ${{N}_{3}}$ is the number of enhanced nodes. The process of generating feature nodes for each window is as follows,

Step 1: Generating a Gaussian random weight matrix of  with dimension of $(f+1)\times {{N}_{1}}$.

Step 2: Putting the value of $w_e$ in the variable of $W_e\left\{i \right\}$, where $i=1\dots N_{2}$, which represents the number of iterations and is equal to the number of generated windows.

Step 3: Letting ${{A}_{1}}=Xw_e$, the random generated weight matrix is used to convolve the features of each sample and obtain new features to form feature nodes of one window.

Step 4: Normalizing ${{A}_{1}}$, then the lasso method is conducted for sparse autoencoder to get a sparse weight matrix of ${{W}_{z}}$, which is given as

\begin{equation}
\begin{aligned}
{{W}_{z}}=\mathrm{argmin} \left(\left\| {{A}_{1}}{{W}_{z}}-X \right\|_{2}^{2}+\lambda {{\left\| {{W}_{z}} \right\|}_{1}} \right),
\end{aligned}
\end{equation}
where $\left\|\cdot\right\|_{2}$ is L2 normalization and $\left\|\cdot\right\|_{1}$ is L1 normalization.

Step 5: Generating feature nodes of ${{Z}_{1}}$ in one window with normalization method, and the feature nodes of ${{Z}_{1}}$ is given as

\begin{equation}
\begin{aligned}
{{Z}_{1}}=\text{norm}(X{{W}_{z}}).
\end{aligned}
\end{equation}

The above process is just the calculation for one window, and for ${N_{2}}$ windows, it will generate ${{N}_{1}}$ feature nodes for every window. Therefore, the matrix of feature node for the whole network is given as

\begin{equation}
\begin{aligned}
Z=\left[ {{Z}_{1}}, {{Z}_{2}},\cdots, {{Z}_{n}} \right],
\end{aligned}
\end{equation}
     
\noindent where $Z$ is the matrix with $s\times ({{N}_{2}}\times {{N}_{1}})$ dimensions, and $n={{N}_{2}}\times {{N}_{1}}$.

\subsubsection{Enhancement nodes}

One of characters for BLS is the enhancement node which is introduced to supplement the feature nodes as the input of the network. Adding enhanced nodes can increase the nonlinear features of the network and better fit the nonlinear function. The process of obtaining enhancement nodes is described as follows,

Step 1: The feature node matrix is standardized and augmented to obtain the enhancement nodes $H$. The weight matrix of ${{W}_{h}}$ is an orthogonal normalized random matrix. The purpose of this operation is to map the feature nodes with linear characteristics to another high-dimensional space, in which way the network has stronger expression ability.

Step 2: The enhancement nodes are activated to achieve the parameter of ${H}'$, which is calculated as

\begin{equation}
\begin{aligned}
{H}'=\mathrm{tansig} \left ( \frac{H {{W}_{h}} s}{\max(H{{W}_{h}})} \right),
\end{aligned}
\end{equation}
where $\mathrm{tansig}(\cdot )$ is a hyperbolic tangent activation function, and $s$ is the zoom factor.

Step 3: The feature nodes and enhancement nodes are combined as input of the entire network, which is given as

\begin{equation}
\begin{aligned}
A=\left[ Z,{H}' \right],
\end{aligned}
\end{equation}

\noindent and the dimension of the input is ${{N}_{2}}\times {{N}_{1}}+{{N}_{3}}$.

\subsubsection{Pseudo-inverse and incremental learning}
The purpose of the network is to obtain the connection weight, which is obtained from

\begin{equation}
\begin{aligned}
W={{A}^{-1}}Y.
\end{aligned}
\end{equation}
It can be calculated by ridge regression as 
\begin{equation}
\begin{aligned}
\mathrm{argmin} \left( \left\| AW-Y \right\|_{2}^{2}+\lambda \left\| W \right\|_{2}^{2} \right),
\end{aligned}
\end{equation}
then we get
\begin{equation}
\begin{aligned}
W={{\left( \lambda I+A{{A}^{T}} \right)}^{-1}}{{A}^{T}}Y,
\end{aligned}
\end{equation}

\noindent where, the expression of generalized inverse of $A$ is defined as

\begin{equation}
\begin{aligned}
{{A}^{+}}=\underset{\lambda \to 0}{\mathop{\lim }}\,{{(\lambda I+A{{A}^{T}})}^{-1}}{{A}^{T}}.
\end{aligned}
\end{equation}
Then, the connection weight is achieved through the above calculations.

In practical applications, more and more sensor data will be collected. When the sensor data volume reaches the order of billions, it will be extremely time-consuming to find the pseudo-inverse of such a matrix with hundreds of millions of rows. At the same time, due to the insufficient fitting ability of the model, the dimension of sensor data will be increased, that is, the number of enhanced nodes will be increased. Therefore, incremental learning algorithm is needed, which is also the essence of BLS. 

Incremental learning is a dynamic and progressive updating algorithm. Its advantage lies in that the updated network connection weight matrix can be obtained with only a small amount of computation through the previous calculation results and the obtained new data.

The input of original network is denoted by ${{A}_{n}}$, the new added enhancement nodes is denoted by $a$. The updated input is given as ${{A}_{n+1}}=[{{A}_{n}}|a]$, and the updated weight is denoted as ${{W}_{new}}={{[{{A}_{n}}|a]}^{-1}}Y$. The generalized inverse of ${{A}_{n+1}}$ is defined as

\begin{equation}
\begin{aligned}
{{A}_{n+1}}^{+}={{[{{A}_{n}}|a]}^{+}}=\left[ \begin{matrix}
   A_{n}^{+}-d{{b}^{T}}  \\
   {{b}^{T}}  \\
\end{matrix} \right],
\end{aligned}
\end{equation}
where $Y$ is the label of training set. From this way, the input and the weight of the entire network are obtained. 

After the training process, the connection weight and normalized parameters need to be saved. The dimension of weight for the entire network is ${{N}_{2}}\times {{N}_{1}}+{{N}_{3}}\times k$, where $k$ is the number of classifications.

\subsection{Fuzzy BLS based on Takagi-Sugeno}
The method of FBLS-TS combines the fuzzy logic and neural network, and adjusts the parameters of the antecedent and subsequent fuzzy system to improve the reasoning ability of the fuzzy system.

\begin{figure}[ht]
    \centering
    \includegraphics[width=12cm]{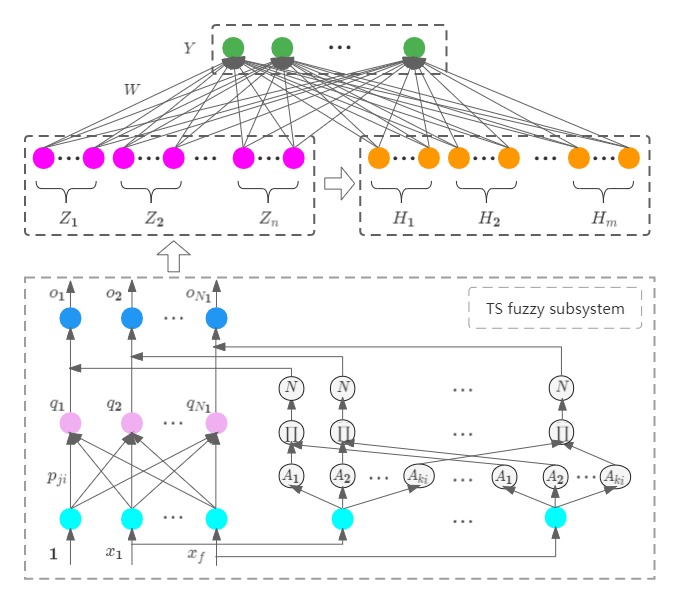}
    \caption{Network structure of the proposed FBLS-TS.}
    \label{BLSTS}
\end{figure}

The BLS model generates feature nodes through randomly generating weights and sparse coefficients, but the FBLS-TS model obtains the feature nodes through the inference and feature extraction from the original input data by Takagi-Sugeno (TS) fuzzy subsystem. The network structure is shown in Figure \ref{BLSTS}.

As seen in the figure, the large dotted box is one adaptive fuzzy subsystem, and it contains two parts which are antecedent network and subsequent network. The tobacco NIR data is denoted as $X={{\left[ {{x}_{1}}, {{x}_{2}}, \cdots , {{x}_{f}} \right]}^{T}}$, and $f$ is the number of feature variables. 

There are four layers in the subsequent network, and the calculation process is as follows,

1) The first layer is the input layer with $f$ neuronal nodes, and the output of this layer is the value of every feature variable.

2) The second layer is the membership function layer, which is used to fuzzify the input data. ${{A}_{1}}, {{A}_{2}}, \ldots, {{A}_{{{N}_{1}}}}$ are fuzzy sets obtained by the input data with the Gaussian function, and they denote the number of the fuzzy rules. The output of this layer is defined as

\begin{equation}
\begin{aligned}
\mu _{i}^{j}={{e}^{-\frac{{{\left( {{x}_{i}}-{{c}_{ij}} \right)}^{2}}}{{{\sigma }_{ij}}^{2}}}}, \qquad{i=1, 2, \cdots, f};  \quad{j=1, 2, \cdots, {k_{i}}},
\end{aligned}
\end{equation}

\noindent where, the center vector of $c$ is obtained from clustering algorithm of K-Mean, and denotes the standard deviation.

3) The third layer is the fuzzy rule layer. Every neuron in this layer represents one fuzzy rule and calculates the fitness. The output of this layer is given as

\begin{equation}
\begin{aligned}
{{w}_{j}}=\mu _{1}^{{{i}_{1}}}\mu _{2}^{{{i}_{2}}}\cdots \mu _{n}^{{{i}_{n}}},
\end{aligned}
\end{equation}
where ${i}_{1} \in \{1,2,\cdots, {{k}_{1}}\}; \, {i}_{2} \in \{1,2,\cdots, \, {{k}_{2}}\}; \cdots; \, {i}_{f} \in \{1,2,\cdots, {{k}_{f}}\}; \, j=1,2,\cdots,k; \, k=\underset{i=1}{\overset{f}{\mathop{\prod }}}\,{{k}_{i}}$.

4) The fourth layer is the normalization layer, and its equation is given as

\begin{equation}
\begin{aligned}
\overline{{{w}_{j}}}=\frac{{{{w}_{j}}}}{{{\sum\limits_{j=1}^{k}{{{w}_{j}}}}}}, \qquad{j=1,2,\cdots ,k}.
\end{aligned}
\end{equation}
 
There are three layers in subsequent network, and the calculation process is as follows, 

1) The first layer is the input layer and its function is to convey the value of feature variable to the next layer.

2) The second layer is the rule layer and every neuronal node represents one rule. Its output is given as

\begin{equation}
\begin{aligned}
{{q}_{j}}={{p}_{j0}}+{{p}_{j1}}{{x}_{1}}+\cdots +{{p}_{jf}}{{x}_{f}}=\sum\limits_{i=0}^{f}{{{p}_{ji}}{{x}_{i}}}.
\end{aligned}
\end{equation}

3) The third layer is the output layer of every fuzzy rule, which is used for generating the feature nodes, and it is calculated as

\begin{equation}
\begin{aligned}
{{o}_{j}}={\overline w}_{j}{{q}_{j}},\text{    } \quad j=1,2,\cdots ,k.
\end{aligned}
\end{equation}

\subsection{Tobacco database}
\begin{figure}[ht]
    \centering
    \includegraphics[width=12cm]{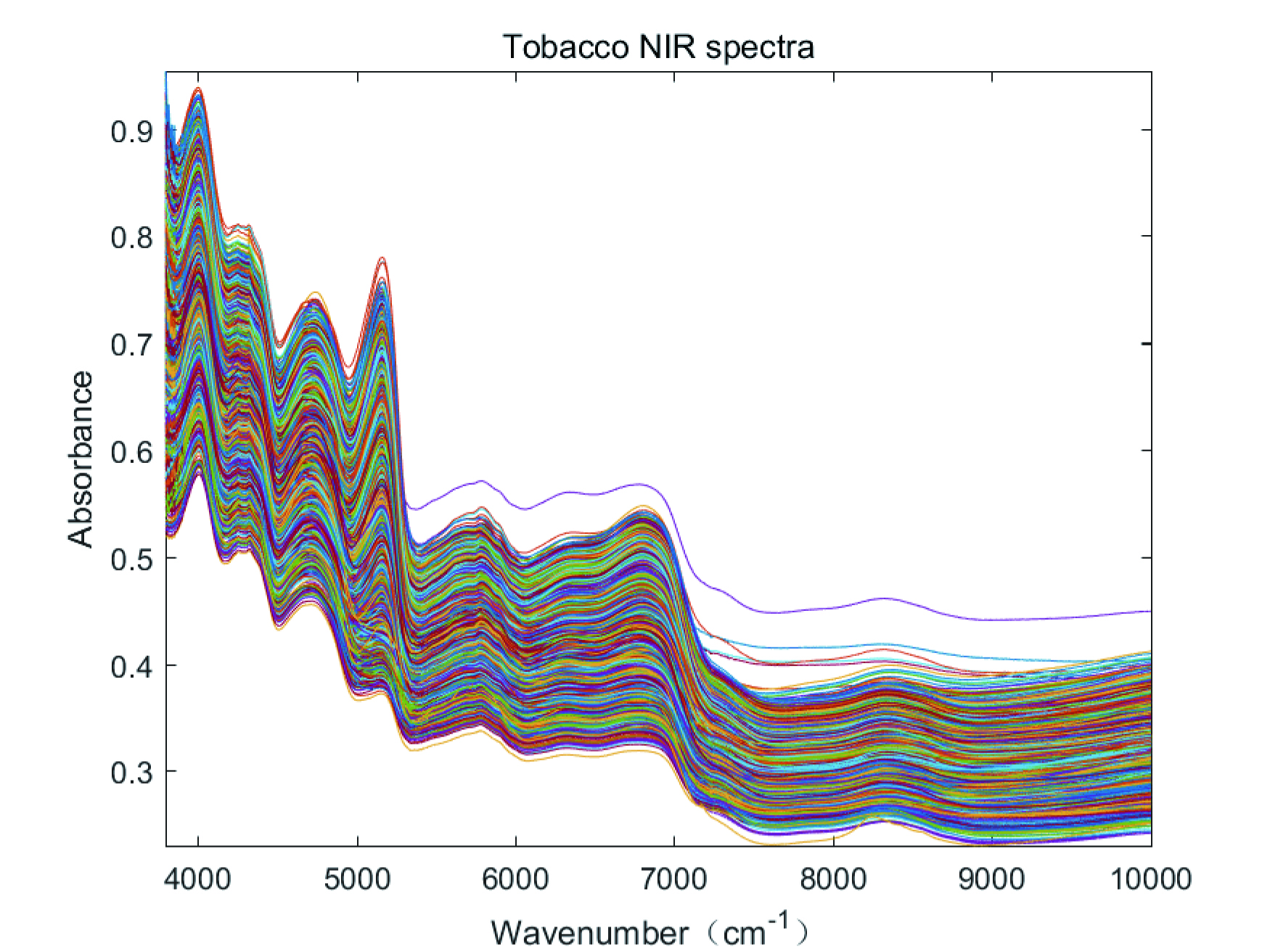}
    \caption{Raw NIR spectra of the 13370 samples.}
    \label{F3}
\end{figure}

A total number of 13370 tobacco samples were collected from eight different areas of Guizhou Province by Guiyang flue-cured tobacco factory in 2018. The tobacco NIR spectra of the data were recorded by Thermo Antaris 2 (Thermo Fisher Scientific Inc. Waltham, USA.) with the spectral resolution of 8 cm$^{-1}$ and 64 scans in the NIR range of 3800 cm$^{-1}$ to 10000 cm$^{-1}$, which is shown in Figure \ref{F3}.

NIR spectrum is generated when energy transition occurs due to anharmonic vibration of molecules and selective absorption of NIR light in specific frequency bands. It mainly records the absorption of frequency doubling and frequency combination of vibration of hydrogen groups (C-H, O-H, N-H, etc.). The absorption intensity and wavelength of NIR light by the same group are different in different physical and chemical environments, which is the principle of qualitative analysis.

Data were randomly divided into training set and testing set, and the regional distribution of training set is shown in Table 1.
\begin{table}[ht]
\centering
\caption{Regional distribution of training set.}
\begin{tabular}{c|p{4cm}}
\hline  
Regions & Sample number  \\  
\hline 
West	&435\\
Northwest	&6116\\
Northeast	&1137\\
Southeast	&265\\
North	&2325\\
South	&114\\
Middle	&131\\
Southwest	&173\\
In total	&10696\\
\hline 
\end{tabular}
\label{t3}
\end{table}

\subsection{Model evaluation}
The prediction accuracy is the significant parameter to access the overall performance in the classification of the tobacco growing area and is defined as

\begin{equation}
\begin{aligned}
{{P}_{\text{a}}}=\frac{{{n}_{r}}}{{{N}_{t}}},
\end{aligned}
\end{equation}
where ${{n}_{r}}$ is the number of samples that are predicted correctly and ${{N}_{t}}$ is the number of samples that are used for prediction. In this study, the number of training set and testing set is 10696 and 2674, respectively.
\section{Results and Discussion}\label{sec:definition}
\sloppy{}
The number of feature nodes, windows and enhanced nodes were tested and the BLS network structure was established according to the experimental results. The experimental results of incremental learning and the comparative experimental results before and after model optimization were analyzed. The results of the model were compared with that of other traditional algorithms. All simulations were implemented on Windows 10 operating system using Matlab R2014a, which is running on a laptop with Intel (R) Core (TM) i5-4210U CPU 1.70 GHz and 2.40 GHz RAM.3.1. 

\subsection{Selection of input node number of network}
\begin{figure}[ht]
    \centering
    \includegraphics[width=12cm]{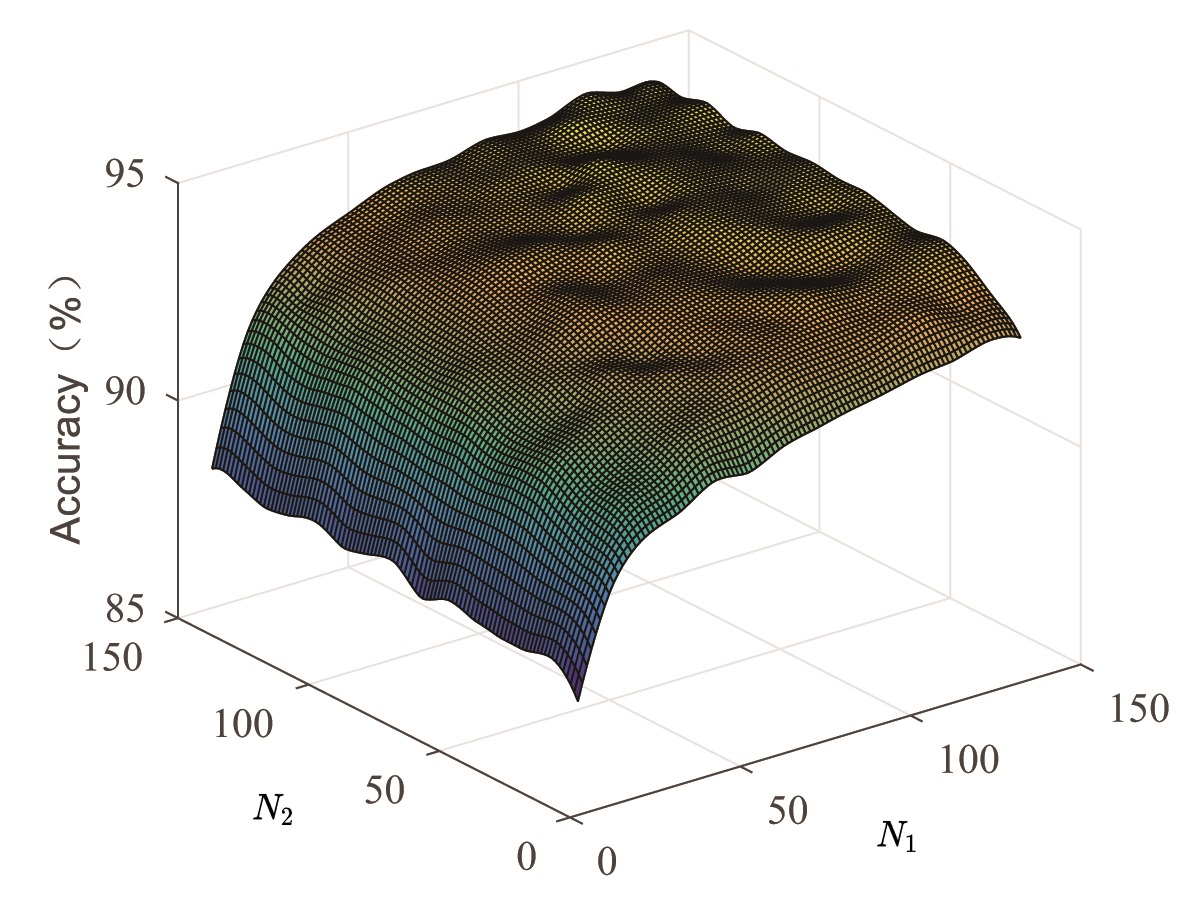}
    \caption{Prediction accuracy of training set along with ${{N}_{1}}$ and ${{N}_{2}}$ (${{N}_{3}}$=1000).}
    \label{F44}
\end{figure}

\begin{figure}[ht]
    \centering
    \includegraphics[width=12cm]{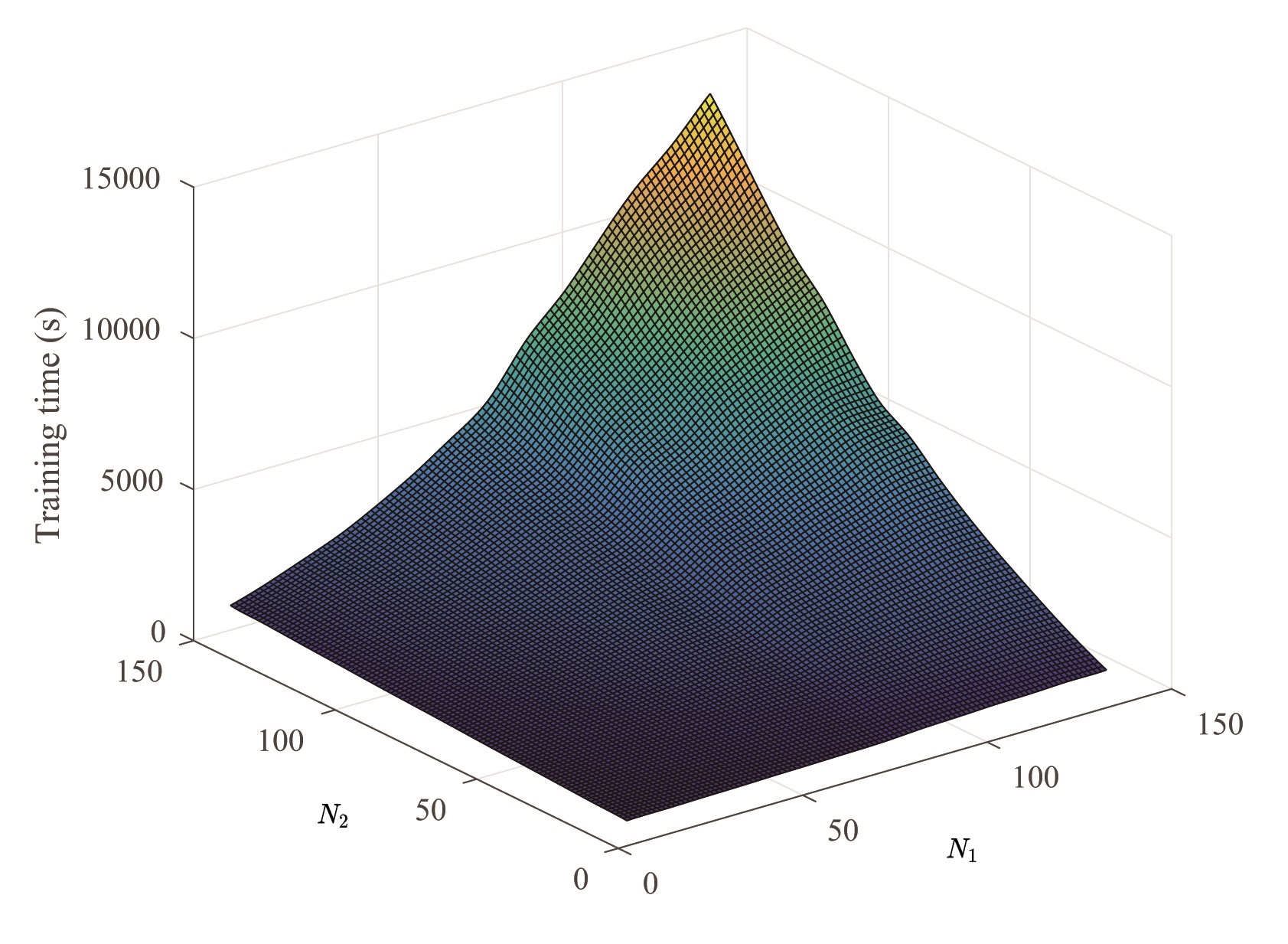}
    \caption{Training time of training set along with ${{N}_{1}}$ and ${{N}_{2}}$ ($N_3$=1000).}
    \label{F5}
\end{figure}

The number of feature nodes, windows and enhancement nodes affects the prediction performance and computing speed of the model. In order to set appropriate parameter values, experiments on the training set and testing set were conducted and results were plotted. The number of windows is denoted by ${{N}_{2}}$, the number of feature nodes in per window is denoted by ${{N}_{1}}$, and the number of enhancement nodes is denoted by ${{N}_{3}}$. When ${{N}_{3}}$ is fixed as 1000, the model is trained at different values of ${{N}_{1}}$ and ${{N}_{2}}$ for training set. The prediction accuracy and training time of the model are shown in Figure \ref{F44} and Figure \ref{F5}, respectively.

It can be seen from Figure 4 that, the prediction accuracy of the model increases along with the increase of the number of windows and feature nodes. It increases rapidly (from 0.85 to 0.90) when the values of ${{N}_{1}}$ and ${{N}_{2}}$ are relatively small (less than 50). However, it increases slowly (from 0.90 to 0.91) when the values of ${{N}_{1}}$ and ${{N}_{2}}$ are relatively large (more than 50). The prediction accuracy achieves as high as 0.9121 when the two parameters are 140 and 150 respectively.

As seen in Figure 5, the training time of the model increases along with the increase of the parameters. It increases slowly when the values of ${{N}_{1}}$ and ${{N}_{2}}$ are small (less than 110) but increases rapidly when the values of ${{N}_{1}}$ and ${{N}_{2}}$ are relatively large (more than 110). The training time gets as high as 131.76 seconds when the two parameters are 140 and 150, respectively.

Set ${{N}_{1}}$=20 and ${{N}_{2}}$=10, the training set was trained with different number of enhancement nodes, and then the model was tested with the testing set. The prediction accuracy and training time obtained were shown in Figure 6. For comparison, the experimental results were plotted while the parameters were set as ${{N}_{1}}$=140 and ${{N}_{3}}$=1000, which is shown in Figure 7.
\begin{figure}[ht]
    \centering
    \includegraphics[width=12cm]{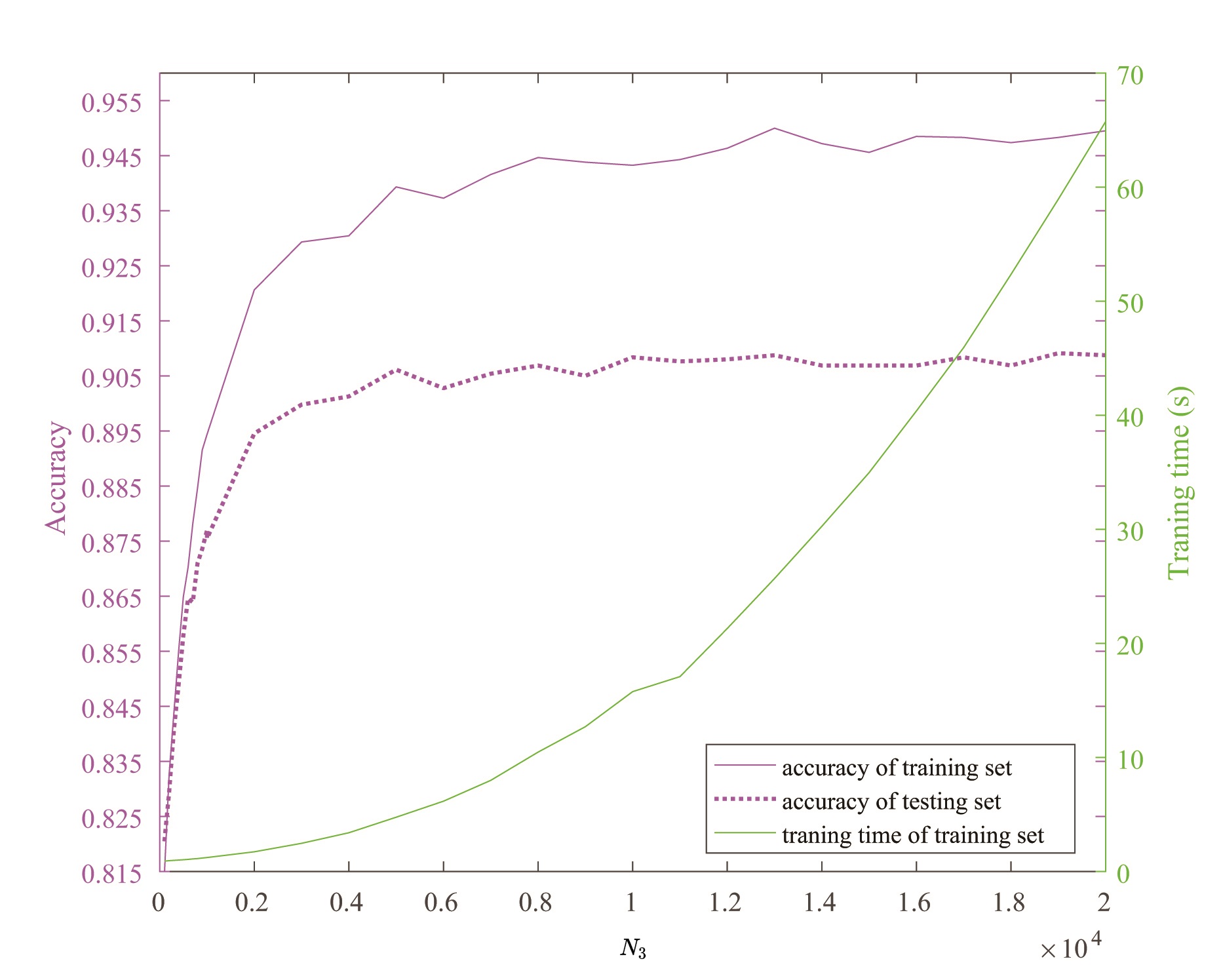}
    \caption{Prediction accuracy and training time of dataset along with ${{N}_{3}}$ (${{N}_{1}}$=20, ${{N}_{2}}$=10).}
    \label{F6}
\end{figure}
\begin{figure}[ht]
    \centering
    \includegraphics[width=12cm]{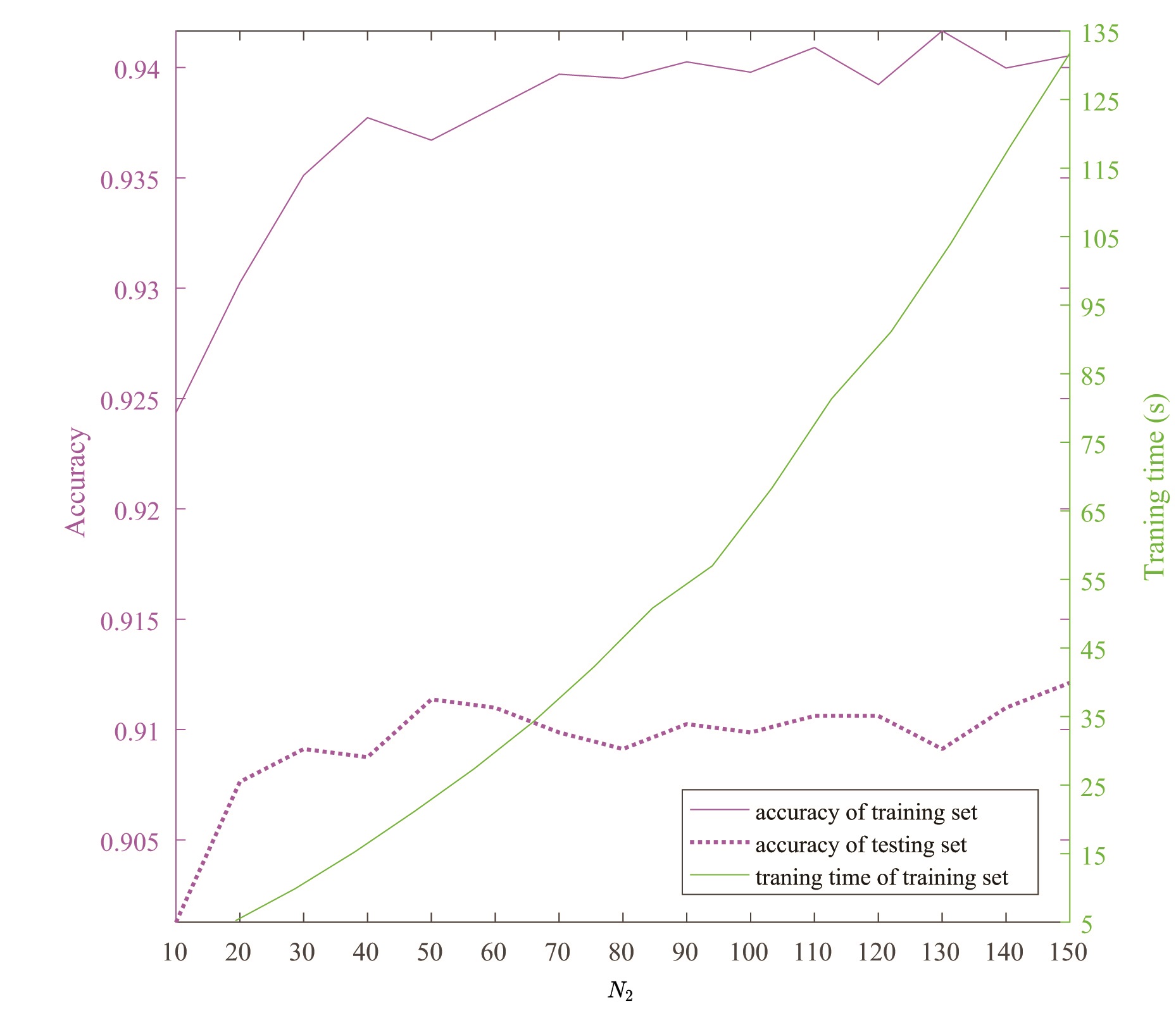}
    \caption{Prediction accuracy and training time of dataset along with ${{N}_{2}}$ ($N_1$=140, $N_3$=1000).}
    \label{F7}
\end{figure}

As seen in Figure \ref{F6}, the curves of prediction accuracy for training set and testing set increase rapidly (from 0.815 to 0.921 and from 0.820 to 0.895) when ${{N}_{3}}$ increases from 100 to 2000, then increase slowly along with the increase of ${{N}_{3}}$, and achieves the highest value with 0.9499 and 0.9088 when ${{N}_{3}}$ is equal to 13000. However, the curve of training time shows the trend of exponential growth along with the increase of ${{N}_{3}}$, and the training time is as high as 25.69 seconds when ${{N}_{3}}$ is equal to 13000.

As seen in Figure \ref{F7}, the curves of prediction accuracy keep the trend of gradually increase (from 0.924 to 0.941 and from 0.901 to 0.912) when ${{N}_{2}}$ increases from 10 to 150, and the curve of training time for training set shows the trend of exponential increase (from 5.2 seconds to 131 seconds).

By comparing the results of Figure 6 and Figure 7, it can be seen that the training time at the highest prediction accuracy obtained in Figure 6 is 25.69 seconds, while it reaches as high as 103.98 seconds at the highest prediction accuracy obtained in Figure 7, which is 78.29 seconds higher than the former. The reason lies in that the feature nodes need to be generated by sparse method and windows need to be iterated at the same time, which are time-consuming, while the generation of enhanced nodes only needs orthogonal normalization iteration, which is far less time-consuming than the generation of feature nodes.

Based on the above analysis, the number of windows in the network is set as 10, the number of feature nodes in each window is set as 20, and the number of enhanced nodes is set as 13000.

\subsection{Experimental results and analysis of incremental learning}
The incremental learning method can be used to analyze the data when the sample size of the data or the dimension of the data increases. After determining the network structure, incremental learning is conducted by increasing the number of feature nodes and enhancement nodes in the network. The experimental results are shown in Table 2.

\begin{table}[ht]
\setlength \tabcolsep{1pt}
\centering
\caption{Step results of dataset classification with incremental learning method.}
\begin{tabular}{cccccc}
\hline  
Number of &Number of &$P_a$ of &$P_a$ of &Extra step &Total training   \\  
 feature nods &enhancement nods &training set &testing set &time (s) & time (s)\\
\hline 
200	&13000	&0.94353&	0.90464	&11.529	&11.529\\
200→210&	13000→14000&	0.9469&	0.90726	&2.4402	&13.9692\\
210→220	&14000→15000&	0.94979	&0.90814&	2.7496&	16.7188\\
220→230	&15000→16000&	0.95409	&0.91090&	2.9349&	19.6537\\
230→240	&16000→17000&	0.95559	&0.91281	&3.2067&	22.8604\\
\hline 
\end{tabular}
\label{t2}
\end{table}

As seen from Table 2, with the increase of the number of feature nodes and enhancement nodes each time, the prediction accuracy of the model for training set and testing set improves (from 0.94353 to 0.95559, and from 0.90464 to 0.91281, respectively). The training time is 11.529 seconds before adding nodes, and the additional one-step training time after adding nodes is very small. When 10 feature nodes and 1000 enhancement nodes are added each time, the additional one-step training time increases from 2.4402 seconds to 3.2067 seconds during four times of incremental learning, and the cumulative training time is only 22.8604 seconds. The reason lies in that ridge regression method adopted in the first calculation of node weights, which may take some time to iterate continuously. However, incremental learning does not need to retrain the model, and its calculation only involves the calculation of matrix multiplication of newly added nodes, so the extra one-step training time is short.

\subsection{Tobacco comparative experimental results of BLS and FBLS}

The BP algorithm is adopted to update the weight of network in the BLS method in this paper, and two incremental approaches are applied, so three modes are used for simulation, which are non-incremental learning, increasing the number of enhancement nodes, and increasing the number of both enhancement nodes and feature nodes. The experimental results are shown in Table 3. In the table, BLS\_BP represents the BLS method that adopts BP algorithm without incremental learning, BLS\_EN represents the BLS method that increases the number of enhancement nodes, and BLS\_EF represents the BLS method that simultaneously increases the number of enhancement nodes and feature nodes. 

As seen from the Table 3, for all the three approaches, the prediction accuracy of FBLS method is always higher than that of the BLS method, which indicates that the feature extraction of TS fuzzy subsystem can effectively improve the prediction performance of the model. However, the training time of FBLS method is always larger than that of BLS method. For example, the training time of FBLS\_EN and BLS\_EN methods is 128.3030 seconds and 50.8731 seconds respectively. The reason lies in that there are more parameters used for feature extraction in TS system, and the update of weight process takes longer time. Therefore, FBLS sacrifices longer time for higher prediction accuracy.
\begin{table}[ht]
\centering
\caption{Comparison of experimental results of BLS and FBLS-TS.}
\begin{tabular}{ccccc}
\hline  
Method&	$P_a$ of & $P_a$ of & Training time &Testing time\\  
 &training set& testing set& (s)& (s)\\
\hline 
BLS\_BP&	0.8548&	0.8512&	57.3322&	0.2156\\
FBLS\_BP	&0.8840	&0.8792&	167.2500&	0.6922\\
BLS\_EN&	0.9514&	0.9061&	50.8731&	0.6044\\
FBLS\_EN	&0.9872&	0.9559	&128.3030&	0.7028\\
BLS\_EF	&0.9413	&0.9046	&91.1280&	0.9171\\
FBLS\_EF	&0.9854&	0.9540&	187. 8224&	0.7508\\

\hline 
\end{tabular}
\label{t3}
\end{table}

\subsection{Advantages of the model in training time}
BLS and FBLS-TS methods are compared with traditional classification algorithms (SVM and GA-SVM), artificial neural network (ANN) and convolution neural network (CNN) in terms of prediction accuracy and training time. The experimental results are shown in Table 4.

\begin{table}[ht]
\centering
\caption{Comparison of classification results with different models.}
\begin{tabular}{ccc}
\hline  
Models&	$P_a$ & Training time (s)
  \\  
\hline 
SVM	&0.8637	&3068.628\\
GA-SVM&	0.9068&	4122.037\\
ANN	&0.8534&	4091.184\\
CNN	&0.9316&	6833.06\\
BLS	&0.9061&	50.8731\\
FBLS-TS	&    0.9559	 &  128.303\\

\hline 
\end{tabular}
\label{t4}
\end{table}

It can be seen from Table 4 that the prediction accuracy of BLS is 0.9061, which is 0.0424 and 0.0527 higher than SVM and ANN, respectively, and it is about the same as that of GA-SVM, which proves that the prediction performance of BLS is superior to the traditional classification algorithms and ANN. However, the training time of BLS is just 50.8731 seconds, but the training time of the other three algorithms are all over 3068 seconds, which shows that BLS has a great advantage in training speed, which is obviously superior to the existing traditional classification algorithms. Compared with CNN, although BLS is 0.0255 lower than CNN in terms of prediction accuracy, its training time is 137 times shorter than CNN. It lies in that CNN is a deep network structure with a large number of hyper-parameters, which makes the training process extremely time$-$consuming. However, BLS has no deep network structure, and expands in the direction of network width with few parameters, and it updates the model through incremental learning, which saves a lot of time for training process. 

The prediction accuracy of the FBLS-TS model is improved to 0.9559, which is 0.0498 higher than that of BLS, and its training time is 128.303 seconds with 77 seconds longer than that of BLS. This is because FBLS-TS model achieves higher prediction accuracy by sacrificing training time.

According to the above analysis, the BLS method before and after improvement has a great advantage over other methods in terms of training time, especially the improved FBLS-TS method achieves better effect than other algorithms in terms of both recognition rate and training speed.

\section{Conclusions}\label{sec:conclusion}
\sloppy{}
The FBLS-TS model is proposed to discriminate the tobacco growing area based on the NIR spectra in this study. The experimental results demonstrate that the FBLS-TS model not only achieves highest prediction accuracy but also has a great advantage in training time, which can address the time$-$consuming problems happened in deep network during training process for big data and retraining the network due to adding more new data. It lies in that the TS fuzzy subsystem adaptively extracted the effective features from data which greatly improves the prediction accuracy, and it adopts the pseudo-inverse calculation and incremental learning method which greatly speed up the training speed of the model.

The performance of the FBLS-TS model is superior to the other traditional algorithms on the classification issue of tobacco growing area. This study not only can be used for classification problems in tobacco industry but also can be extended for application in agriculture, medicine, food, petrochemical, environment and other fields.

\section*{Acknowlegments}
This research was funded by the Key Program for Science and Technology of China National Tobacco Corporation (Grant No. 110202102038), National Natural Science Foundation of China (Grant No. 62103068; Grant No. 51978111), and Chongqing Municipal Education Commission (Grant No. KJQN202100745).

\bibliography{mybibfile.bib}

\begin{thebibliography}{10}
\expandafter\ifx\csname url\endcsname\relax
  \def\url#1{\texttt{#1}}\fi
\expandafter\ifx\csname urlprefix\endcsname\relax\def\urlprefix{URL }\fi
\expandafter\ifx\csname href\endcsname\relax
  \def\href#1#2{#2} \def\path#1{#1}\fi

\bibitem{wd_2019_lar}
D.~Wang, F.~Tian, Z.~Zhu, W.~Pan, {Automatic Prediction of Leave Chemical
  Compositions Based on Nir Spectroscopy with Machine Learning}, International
  Journal of Robotics and Automation 34~(4) (2019) 391--396.

\bibitem{wd_2017}
D.~Wang, F.~Tian, S.~X. Yang, Z.~Zhu, {Intelligent Estimate of Chemical
  Compositions Based on NIR Spectra Analysis}, in: 2017 IEEE International
  Conference on Information and Automation (ICIA), Macau, China, 2017, pp.
  472--477.

\bibitem{tobacco_1}
M.~S. Rahman, N.~F. Ahmed, M.~Ali, M.~M. Abedin, M.~S. Islam, {Determinants of
  Tobacco Cultivation in Bangladesh}, Tobacco Control 29~(6) (2020) 692--694.

\bibitem{tobacco_2}
H.~Liu, Y.~Zhang, X.~Zhou, X.~You, Y.~Shi, J.~Xu, {Source Identification and
  Spatial Distribution of Heavy Metals in Tobacco-growing Soils in Shandong
  Province of China with Multivariate and Geostatistical Analysis},
  Environmental Science and Pollution Research 24~(6) (2017) 5964--5975.

\bibitem{tobacco_3}
H.~Wu, Q.~Liu, J.~Ma, L.~Liu, Y.~Qu, Y.~Gong, S.~Yang, T.~Luo, {Heavy Metal
  (loids) in Typical Chinese Tobacco-growing Soils: Concentrations, Influence
  Factors and Potential Health Risks}, Chemosphere 245 (2020) 125591.

\bibitem{c1}
B.~Chen, W.~Xing, D.~Lu, {Application of Tobacco Leaf Origin Identification
  with X - ray Fluorescent Based on Discriminant Analysis}, Journal of Jiangsu
  University (Natural Science edition) 5 (2015) 545--549.

\bibitem{c2}
X.~Zhu, Y.~Zong, Y.~Li, J.~Xie, {Identification of Flue-cured Tobacco from
  Different Countries with Electronic Noses}, Tobacco Science \& Technology 3
  (2008).

\bibitem{c3}
S.~Williams, S.~Hubbard, K.~Reinhard, S.~Chaves, {Establishing Tobacco Origin
  from Pollen Identification: An Approach to Resolving the Debate}, Journal of
  Forensic Sciences 59 (2014) 1642--1649.

\bibitem{c4}
S.~Wu, T.~Liu, J.~Ge, Y.~Sha, {Chemical Composition - Naive Bayesian
  Classification Algorithm for Tobacco Leaf Origin Pattern Recognition},
  Journal of Henan Normal University (Natural Science Edition) 01 (2018).

\bibitem{c5}
L.~Wang, J.~Zang, Q.~Zhang, Z.~Niu, G.~Hua, N.~Zheng, {Action Recognition by an
  Attention-Aware Temporal Weighted Convolutional Neural Network}, Sensors 18
  (2018) 1979.

\bibitem{c6}
B.~Richter, M.~Rurik, S.~Gurk, O.~Kohlbacher, M.~Fischer, {Food Monitoring:
  Screening of the Geographical Origin of White Asparagus Using FT-NIR and
  Machine Learning}, Food Control 104 (2019) 318--325.

\bibitem{c7}
T.~Wu, I.~Tung, H.~Hsu, C.~Kuo, J.~Chang, S.~Chen, C.~Tsai, Y.~Chuang,
  {Quantitative Analysis and Discrimination of Partially Fermented Teas from
  Different Origins Using Visible/Near-Infrared Spectroscopy Coupled with
  Chemometrics}, Sensors 20 (2020) 5451.

\bibitem{c8li2020near}
H.~Li, D.~Jiang, J.Cao, D.~Zhang, {Near-Infrared Spectroscopy Coupled
  Chemometric Algorithms for Rapid Origin Identification and Lipid Content
  Detection of Pinus Koraiensis Seeds}, Sensors 20 (2020) 4905.

\bibitem{c9}
Q.~Xiao, X.~Bai, P.~Gao, Y.~He, {Application of Convolutional Neural
  Network-Based Feature Extraction and Data Fusion for Geographical Origin
  Identification of Radix Astragali by Visible/Short-Wave Near-Infrared and
  Near Infrared Hyperspectral Imaging}, Sensors 20 (2020) 4940.

\bibitem{c10liu2019rapid}
F.~Liu, W.~Wang, T.~Shen, J.~Peng, W.~Kong, {Rapid Identification of Kudzu
  Powder of Different Origins Using Laser-induced Breakdown Spectroscopy},
  Sensors 19 (2019) 1453.

\bibitem{wd_svm}
D.Wang, L.~Xie, S.~X. Yang, F.~Tian, {Support Vector Machine Optimized by
  Genetic Algorithm for Data Analysis of Near-Infrared Spectroscopy Sensors},
  Sensors 18~(10) (2018) 3222.

\bibitem{c11}
C.~Zhu, H.~Gong, Z.~Li, C.~Yu, {Application of High Dimensional Feature
  Grouping Method in Near-infrared Spectra of Identification of Tobacco Growing
  Areas}, in: 2016 3rd International Conference on Information Science and
  Control Engineering (ICISCE), Beijing, China, 2016, pp. 230--234.

\bibitem{c12}
Y.~Wang, X.~Ma, Y.~Wen, C.~Yu, L.~Wang, L.~Zhao, J.~Li, {Tobacco Quality
  Analysis of Producing Areas of Yunnan Tobacco Using Near-infrared (nir)
  Spectrum}, Spectroscopy and Spectral Analysis 33 (2013) 78--80.

\bibitem{c13}
S.~Zhou, B.~Tan, {Electrocardiogram Soft Computing Using Hybrid Deep Learning
  CNN-ELM}, Applied Soft Computing 86 (2020) 105778.

\bibitem{c14}
S.~J. Fong, G.~Li, N.~Dey, R.~G. Crespo, E.~Herrera-Viedma, {Composite Monte
  Carlo Decision Making Under High Uncertainty of Novel Coronavirus Epidemic
  Using Hybridized Deep Learning and Fuzzy Rule Induction}, Applied Soft
  Computing 93 (2020) 106282.

\bibitem{c15}
M.~Mittal, L.~M. Goyal, S.~Kaur, I.~Kaur, A.~Verma, D.~J. Hemanth, {Deep
  Learning Based Enhanced Tumor Segmentation Approach for MR Brain Images},
  Applied Soft Computing 78 (2019) 346--354.

\bibitem{c16}
Y.-S. Su, C.-F. Ni, W.-C. Li, I.-H. Lee, C.-P. Lin, {Applying Deep Learning
  Algorithms to Enhance Simulations of Large-scale Groundwater Flow in IoTs},
  Applied Soft Computing 92 (2020) 106298.

\bibitem{c16_1}
S.~Afrasiabi, M.~Afrasiabi, B.~Parang, M.~Mohammadi, {Designing a Composite
  Deep Learning Based Differential Protection Scheme of Power Transformers},
  Applied Soft Computing 87 (2020) 105975.

\bibitem{c16meng2018study}
M.~Lu, K.~Yang, P.~Song, R.~Shu, L.~Wang, Y.~Yang, H.~Liu, J.~Li, L.~Zhao,
  Y.~Zhang, {The Study of Classification Modeling Method for Near Infrared
  Spectroscopy of Tobacco Leaves Based on Convolution Neural Network},
  Spectroscopy and Spectral Analysis 38 (2018) 3724--3728.

\bibitem{c17lee2018deep}
M.~Lee, Y.~Xiong, G.~Yu, G.~Li, {Deep Neural Networks for Linear Sum Assignment
  Problems}, IEEE Wireless Communications Letters 7 (2018) 962--965.

\bibitem{c18wang2020improved}
D.~Wang, F.~Tian, S.~Yang, Z.~Zhu, D.~Jiang, B.~Cai, {Improved Deep CNN with
  Parameter Initialization for Data Analysis of Near-Infrared Spectroscopy
  Sensors}, Sensors 20 (2020) 874.

\bibitem{c19chen2017broad}
C.~Chen, Z.~Liu, {Broad Learning System: An Effective and Efficient Incremental
  Learning System without the Need for Deep Architecture}, IEEE Transactions on
  Neural Networks and Learning Systems 29 (2017) 10--24.

\bibitem{c20chen2018universal}
C.~Chen, Z.~Liu, S.~Feng, {Universal Approximation Capability of Broad Learning
  System and Its Structural Variations}, IEEE Transactions on Neural Networks
  and Learning Systems 30 (2018) 1191--1204.

\bibitem{c21luo2018intelligent}
S.~Luo, C.~Zhao, Y.~Fu, {An Intelligent Human Activity Recognition Method with
  Incremental Learning Capability for Bedridden Patients}, in: 2018 15th
  International Conference on Control, Automation, Robotics and Vision
  (ICARCV), IEEE, 2018, pp. 1284--1289.

\end{thebibliography}

\end{document}